\documentclass[lettersize,journal]{IEEEtran}
\usepackage{lineno}
\usepackage{amsmath,amsfonts}
\usepackage{array}
\usepackage{tabularray}
\usepackage[caption=false,font=normalsize,labelfont=sf,textfont=sf]{subfig}
\usepackage{textcomp}
\usepackage{stfloats}
\usepackage{url}
\usepackage{verbatim}
\usepackage{graphicx}
\usepackage{adjustbox}
\usepackage{xcolor}
\usepackage{pifont}
\usepackage{amssymb}
\usepackage{multirow} 
\newcommand{\cmark}{\ding{51}} 

\usepackage[ruled, lined, linesnumbered, commentsnumbered, longend]{algorithm2e}       
\usepackage[pagebackref=false,breaklinks=true,letterpaper=true,colorlinks,bookmarks=false,citecolor=citecolor,linkcolor=linkcolor]{hyperref}
\definecolor{citecolor}{HTML}{0071BC}
\definecolor{linkcolor}{HTML}{ED1C24}

\begin{document}
\title{A Gated Cross-domain Collaborative Network for Underwater Object Detection}
\author{Linhui Dai, Hong Liu$^*$ \IEEEmembership{Member, IEEE}, Pinhao Song, Mengyuan Liu
\thanks{Linhui Dai, Hong Liu, and Mengyuan Liu are with the Key Laboratory of Machine Perception, Peking University, Shenzhen Graduate School, Shenzhen, China (Email: dailinhui@pku.edu.cn, hongliu@pku.edu.cn, nkliuyifang@gmail.com).} 
\thanks{Pinhao Song is with the Robotics Research Group, KU Leuven, Leuven,
Belgium (Email: pinhao.song@kuleuven.be).} }
\maketitle
\begin{abstract}
Underwater object detection (UOD) plays a significant role in aquaculture and marine environmental protection. Considering the challenges posed by low contrast and low-light conditions in underwater environments, several underwater image enhancement (UIE) methods have been proposed to improve the quality of underwater images. However, only using the enhanced images does not improve the performance of UOD, since it may unavoidably remove or alter critical patterns and details of underwater objects. In contrast, we believe that exploring the complementary information from the two domains is beneficial for UOD. The raw image preserves the natural characteristics of the scene and texture information of the objects, while the enhanced image improves the visibility of underwater objects. Based on this perspective, we propose a Gated Cross-domain Collaborative Network (GCC-Net) to address the challenges of poor visibility and low contrast in underwater environments, which comprises three dedicated components. Firstly, a real-time UIE method is employed to generate enhanced images, which can improve the visibility of objects in low-contrast areas. Secondly, a cross-domain feature interaction module is introduced to facilitate the interaction and mine complementary information between raw and enhanced image features. Thirdly, to prevent the contamination of unreliable generated results, a gated feature fusion module is proposed to adaptively control the fusion ratio of cross-domain information. Our method presents a new UOD paradigm from the perspective of cross-domain information interaction and fusion. Experimental results demonstrate that the proposed GCC-Net achieves state-of-the-art performance on four underwater datasets. 
\end{abstract}

\section{Introduction}
The ocean is one of the most important natural resources on Earth, playing a crucial role in the balance of the planet's climate, biodiversity, and economy. The underwater environment is a vast and complex ecosystem, characterized by unique physical and biological features that are challenging to study and monitor \cite{li2021underwater, chen2020reveal,dai2023edgeguided}. Currently, exploration of underwater environments relies on various types of equipment, including sonar, visually-guided autonomous underwater vehicles (AUVs), and remotely operated vehicles (ROVs). By being equipped with underwater cameras, AUVs and ROVs can detect and observe underwater objects \cite{islam2020fast, EfficientDets}. For an AUV or RUV to perform object perception, underwater object detection is a vital
computer vision task aiming to locate and recognize marine organisms or marine debris. 

\begin{figure*}[!t]
  \centering
    \subfloat[Preprocessing]{\includegraphics[width=0.45\linewidth, keepaspectratio]{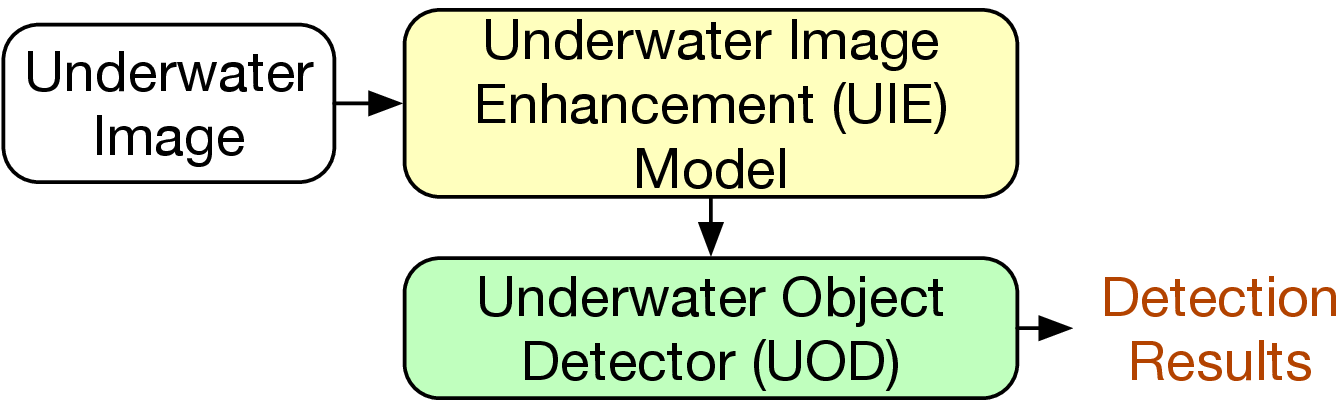}\label{fig1a}}
    \subfloat[Multi-task Learning]{\includegraphics[width=0.47\linewidth, keepaspectratio]{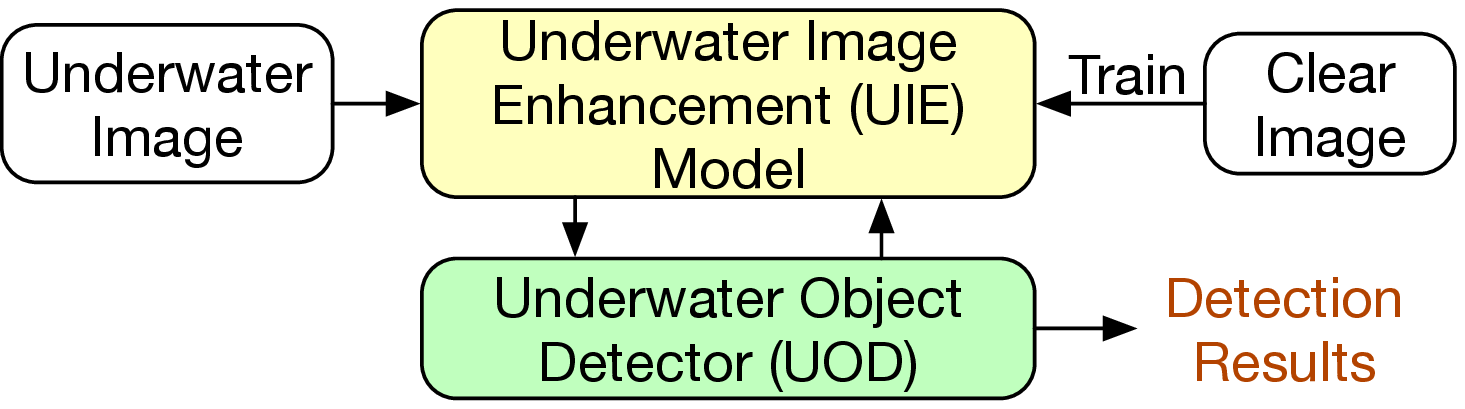}\label{fig1b}}\\
    % \vspace{0.5cm}
    \subfloat[Cross-domain Collaboration (ours)]{\includegraphics[width=0.60\linewidth, keepaspectratio]{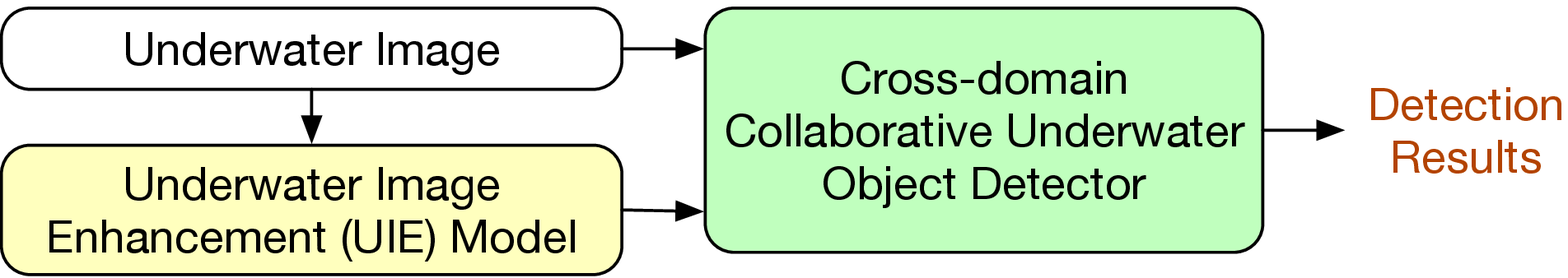}\label{fig1c}}
     \caption{Comparisons of different application manners of underwater image enhancement (UIE) models in underwater object detection (UOD) task. (a): For the preprocessing manner, the UIE method is employed as a preprocessing task, e.g., Reveal \cite{chen2020reveal}. (b): Some methods treat joint training of the UIE and UOD methods as a multi-task learning problem, optimizing both models simultaneously, e.g., TACL \cite{liu2022twin}. (c): Our method presents a new cross-domain collaborative paradigm to explore the interaction and fusion between the image features from the two domains.}
     \label{fig:intro}
  \end{figure*}

The biggest difference between underwater object detection (UOD) and generic object detection (GOD) \cite{zhu2020autoassign,qiao2021detectors, dai2022ao2} lies in the different detection environments. The underwater images inevitably suffer from poor visibility, light refraction, absorption, and scattering \cite{islam2020fast}. Several works \cite{fu2022uncertainty, dong2022underwater, li2021underwater} have focused on improving the quality of underwater images, and they all indicate that images processed through underwater image enhancement (UIE) are helpful for subsequent object detection tasks. However, some works \cite{liu2022twin, chen2020reveal} have pointed out that only using enhanced images may not benefit the effectiveness of detection and even lead to a severe performance drop. This is because although these UIE methods achieve better visual restoration in human perception, algorithms perceive scenes differently from human eyes. Simply correcting the color and contrast of the underwater scene does not necessarily facilitate the understanding of the scene. Additionally, the enhancement process may unavoidably remove or alter critical patterns and details, which may introduce noise or artifacts into the image \cite{liu2022twin, chen2020reveal}. Therefore, there have been some works \cite{liu2022twin, chen2020reveal, yeh2021lightweight} exploring the relationship between underwater visual enhancement and object detection. These methods could be roughly categorized into two different manners: (1) preprocessing manner and (2) multi-task learning manner, as shown in Fig. \ref{fig:intro}. In the preprocessing manner, the UIE method is employed as a preprocessing step to generate enhanced images, as shown in Fig. \ref{fig1a}. However, the effectiveness of detection algorithms cannot be significantly improved because the UIE methods may introduce
noise or artifacts that hampers the detection process. In the multi-task learning manner, some works \cite{liu2022twin, sun2022rethinking} optimize visual restoration and object detection as a multi-task joint learning problem to guide the enhancement towards the detection-pleasing direction, as shown in Fig. \ref{fig1b}. However, these methods require matched pairs of degraded and clear images to train UIE models and it is difficult to obtain the paired images in practice.

Unlike previous methods, we present a novel cross-domain collaborative paradigm that involves the interaction and fusion of features from the UIE-generated enhanced image domain and the raw image domain, as shown in Fig. \ref{fig1c}. Both the UIE-generated enhanced image domain and the raw image domain are beneficial for underwater object detection for each domain provides unique and valuable information. Specifically, the enhanced images help to improve the visual quality of underwater images, providing clearer and more discernible object boundaries, which is beneficial in addressing issues of low contrast and low-light encountered in UOD. On the other hand, the raw images preserve the natural characteristics of the scene and provide clear texture information about objects. We argue that exploring the interaction and fusion between the raw and enhanced images will be beneficial for accurately identifying underwater objects. There are two major challenges in the process of collaboration: (1) how to efficiently get the promising enhanced images and then extract the enhanced features? (2) how to interact and fuse with the features extracted from the enhanced images and raw images?

In this paper, we propose a Gated Cross-domain Collaborative Network (GCC-Net) that learns the interaction and fusion between the image features from the two domains. Our method provides a novel manner to integrate UIE and UOD in a unified framework, as shown in Fig. \ref{fig1c}. The proposed unified framework has three dedicated components to address the above challenges. Specifically,  \textbf{for the challenge (1)}, it is difficult to retrain new UIE models for the UOD task due to the lack of abundant paired clear and distorted underwater images and the inability of synthetic data to fully simulate all underwater conditions. Therefore, we develop a real-time UIE model named ``water-MSR'' based on the existing multi-scale Retinex model \cite{land1964retinex} and IMSRCP \cite{tang2019efficient}. The proposed model accelerates the base model by progressively reducing the image size and recursively invoking the fast filtering algorithm. It enables real-time integration of the UIE model into the object detection framework. \textbf{For the challenge (2)}, we propose a transformer-based cross-domain feature interaction (CFI) module to capture multi-domain feature interaction and collaboration. In addition, we design a gated feature fusion (GFF) mechanism that uses a gate function to control the fusion rate of the raw and enhanced domain information adaptively. The gated mechanism mitigates the negative impact caused by some unreliable results generated by water-MSR.

Our main contributions can be summarized as follows:
\begin{itemize}
\item We propose a Gated Cross-domain Collaborative network (GCC-Net) to address the challenges of poor visibility and low contrast in underwater environments. Our model simultaneously takes the raw image and the enhanced image in parallel, and then interacts and fuses information from both domains within a single framework.
\item A real-time UIE method water-MSR is developed to generate enhanced images, which can improve the visibility of underwater objects in low-light areas. Simultaneously learning features from both raw and enhanced image domains is beneficial for identifying underwater objects under low-contrast and low-light conditions.
\item A novel cross-domain feature interaction (CFI) module is designed to facilitate feature interaction and explore complementary information between the enhanced and raw images. In addition, a gated feature fusion (GFF) mechanism is proposed to adaptively control the fusion ratio of cross-domain information, thereby avoiding contamination from some unreliable generated results.
\item The extensive experiments on four public underwater datasets: DUO \cite{DUO}, Brackish \cite{brackish}, TrashCan \cite{hong2020trashcan}, and WPBB \cite{EfficientDets} demonstrate the effectiveness of the proposed model. The proposed GCC-Net achieves state-of-the-art performance on these four datasets. Our code will be released at \url{https://github.com/Ixiaohuihuihui/GCC-Net}.

\end{itemize}

\section{Related Works}

\subsection{Underwater Object Detection}
Over the past few years, underwater object detection has garnered much research attention considering its wide applications in marine engineering and aquatic robotics. Currently, the most widely used method in underwater object detection is based on deep learning. Since the existing underwater datasets are limited in quantity, some works \cite{ROIMIX, huang2019faster, liu2020towards, chen2023achieving, UDD} aim to increase the diversity of the data by data augmentation. For instance, Lin et al. \cite{ROIMIX} propose an augmentation method called ROIMIX to conduct proposal-level fusion among multiple images. Poisson GAN \cite{UDD} is also a data augmentation method, which uses Poisson blending to change the number, position, and even size of objects in an image. Several works \cite{chen2022swipenet, song2023boosting} improve the detection performance by raising the capability of feature extraction. For example, Chen et al. \cite{chen2022swipenet} propose SWIPENET which consists of high-resolution and semantic rich hyper feature maps. Song et al. \cite{song2023boosting} propose a two-stage detector Boosting R-CNN which can provide high-quality proposals and consider objectness and IoU prediction for uncertainty to model the object prior probability. The UOD models need to be mounted on battery-powered AUVs. Thus, the detection model should be lightweight and real-time. Yeh et al. \cite{yeh2021lightweight} present a color conversion module to transform color images to the corresponding grayscale images to solve the problem of underwater color absorption, thereby improving the object detection performance with lower computational complexity. Zocco et al. \cite{EfficientDets} propose an efficient detector for marine debris detection by changing the layer distribution and architecture of EfficientDet \cite{tan2020efficientdet}.

Due to the inherent challenges of underwater environments, such as low contrast, various lighting conditions, and changes in water quality, image enhancement is necessary. Existing methods \cite{liu2022twin, chen2020reveal} attempt to use underwater image enhancement as a preprocessing step and perform object detection on the enhanced images. However, the detection performance is unsatisfactory and even leads to a serve performance drop. Since the algorithms perceive the scene differently from human eyes, correcting the color and contrast of underwater images may not necessarily help to understand the scene, and the enhanced image may lose some important details and potential value that are beneficial for object detection. Liu et al. \cite{liu2022twin} address the problem by considering it as a multi-task optimization problem. They propose an object-guided twin adversarial and contrastive learning network to achieve a good balance between visual restoration and detection accuracy. However, this method requires pairs of distorted-clear images to train a UIE model and treat the object detection task as a verification task. In practice, the synthetic training samples cannot completely simulate all underwater conditions, retraining a UIE model is thankless for UOD tasks.

Unlike the above methods, we propose to use the existing UIE methods to improve the performance of UOD. The enhanced image can correct the color and contrast of the underwater scene, while the raw underwater image can provide clear texture information and potential value. Therefore, we propose a unified framework that integrates image enhancement and object detection models into a single framework and improves the performance of underwater object detection by exploring complementary information and achieving interaction and fusion between the two image modalities.

\subsection{Underwater Image Enhancement}
Existing underwater cameras are generally carried on vision-guided AUVs. Although high-quality cameras are used, underwater images are affected by light absorption and scattering in the water. Therefore, the original underwater images and videos rarely meet the expectations of visual quality, which will affect the performance of pattern recognition and object detection. Since deep learning networks are trained by high-quality images, the learning-based algorithms assume that the input is a clear image. The existing UIE models usually use prior knowledge or statistical assumptions about the scene. Chiang et al. \cite{underwatermodel} propose an underwater image formation model to enhance underwater images by a dehazing algorithm. The proposed underwater image degradation model is widely used in various UIE methods. Li et al. \cite{watergan} employ a Generative Adversarial Network (GAN) and an image formation model to synthesize degraded/clean image pairs for reconstructing clear images. Li et al. \cite{li2020underwater} design an underwater image synthesis algorithm that can simulate different underwater types and degradation levels and then reconstruct the clear latent underwater image. Islam et al. \cite{islam2020fast} propose a conditional generative adversarial network-based model for real-time underwater image enhancement. The Sea-thru method \cite{Sea-Thru} is proposed to recover lost colors in underwater images by estimating backscatter. Li et al. \cite{USP} propose an underwater image synthesis algorithm based on underwater scenes prior to simulating a diverse set of degraded underwater images. Though many of them mention their potential value as a pre-processing step for high-level downstream tasks (e.g., object detection and segmentation), as far as we know, no one has applied these methods to UOD tasks. Due to the significant domain shift between the UIE datasets (such as the EUVP dataset \cite{islam2020fast}) and the UOD datasets, and the absence of corresponding clear-scene datasets in the UOD dataset, retraining the UIE model on the UOD dataset seems to be very challenging. 

Unlike treating existing deep learning-based UIE methods as offline preprocessing steps, our method integrates a UIE model and a UOD model simultaneously into a unified framework, allowing the UIE model to assist the UOD model. Our method is the first to improve the performance of underwater object detection from the perspective of cross-domain data interaction and fusion.

\section{Method}
\subsection{Overview}
\begin{figure*}[!t] 	
\centering 	
\includegraphics[width=\textwidth,height=\textheight,keepaspectratio]{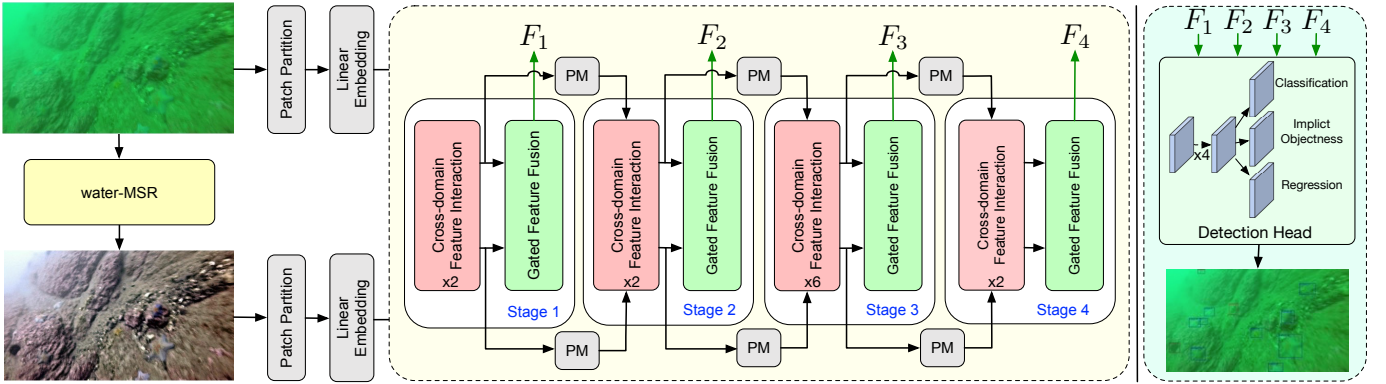}  	
\caption{Illustration of our proposed framework. It mainly consists of four components: a water-MSR module, the cross-domain feature interaction blocks, four gated feature fusion modules, and a detection head.} 	
\label{fig:framework} 
\end{figure*}

The framework of the proposed method is shown in Fig. \ref{fig:framework}, which is mainly composed of four components: (1) an online UIE model water-MSR is used to generate desired enhanced images, (2) the cross-domain feature interaction (CFI) module is proposed to facilitate feature interaction and explore complementary information between the enhanced and raw images, (3) a gated feature fusion (GFF) module is introduced to control the fusion rate of two domains, (4) a detection head is used to output the results.

The proposed method takes an image as input and predicts the labels and positions of objects. Given an image, the online UIE model is first used to generate the enhanced image as the clear domain. To facilitate cross-domain interaction and fusion, we use Swin Transformer \cite{liu2021swin} as the backbone, which constructs hierarchical feature maps and has linear computational complexity to image size. The two images will be split into non-overlapping patches by a patch partition module, respectively. A linear embedding layer is used to project the raw feature into an arbitrary dimension. 

Then, these patch tokens will be sent to the proposed cross-domain feature interaction (CFI) module, followed by a 2-layer MLP with GELU non-linearity in between. Through the proposed CFI module, we can facilitate the information interaction and integration of the clear domain and raw domain. Next, we need to fuse the complementary information of two domains. In order to avoid image area noise caused by unreliable enhanced images, we use the proposed gated feature fusion (GFF) mechanism to control the fusion ratio. In the proposed method, there are four stages, which correspond to four GFF modules. Following the Swin-T \cite{liu2021swin}, the numbers of cross-domain feature interaction modules are $\{2,2,6,2\}$ of the stages, respectively. As shown in Fig. \ref{fig:framework}, the green lines represent the outputs of the GFF module. These four features are then fed into the detection head. Finally, a detection head is used to output the localization and classification results.

\subsection{Online Underwater Image Enhancement Model}
Note that the goal of this paper is not to develop a new UIE method. Due to significant variations in water quality and the difficulty in obtaining paired training samples in practice, retraining a UIE method cannot fulfill the detection requirements for all water conditions. We hope to utilize the existing physical UIE models and integrate them into UOD tasks. Accordingly, we choose multi-scale Retinex \cite{jobson1997multiscale} and IMSRCP \cite{tang2019efficient} as the base UIE method. To embed the UIE model in real-time within the UOD framework, we propose a real-time UIE model ``water-MSR'' by recursively invoking fast filtering. 

Firstly, we briefly review the multi-scale Retinex (MSR) model. The Retinex theory models the imaging process of objects by separating the color imaging process into two components: the illuminant and the reflectance, which can eliminate the influence of incident light to recover the characteristics of the object from noisy images. Multi-scale Retinex algorithm is a variation of Retinex, which can capture light changes at different scales and remove the influence of illumination from input images, achieving dynamic range compression, color consistency, and brightness reproduction simultaneously. Next, we propose to accelerate the base UIE model by replacing the Gaussian filter operation with fast filtering based on recursive image pyramids.  

Given an underwater raw image $I_1$, the online water-MSR is developed to generate the enhanced image $I_2$. The online water-MSR algorithm is mainly composed of two steps. Firstly, the color pre-correction is conducted to equalize the degraded underwater images and reduce the dominant color. Denote $I_{mean}$ and $I_{var}$ as the mean value and standard deviation of the raw image $I_1$, then the minimum $I_{min}$ and maximum $I_{max}$ of each RGB channel are extracted by:
\begin{equation}
\label{eq1}
  \begin{aligned}
&I_{\text {min }}^i=I_{\text {mean }}^i- I_{\mathrm{var}}^i, \\
&I_{\mathrm{max}}^i=I_{\text {mean }}^i+ I_{\mathrm{var}}^i, \\
&I^i=\frac{I^i-I_{\text {min }}^i}{I_{\max }^i-I_{\min }^i} \times 255,
\end{aligned}
\end{equation}
where $i \in\{R, G, B\}$ is the color channel of the RGB image and $I$ is the color pre-corrected image.

Secondly, we employ a weighted fusion of multi-scale Retinex model to balance dynamic range compression and color rendition. To accelerate the algorithm speed, we adopt the fast Gaussian filter by following a recursive approach, and we can progressively downsample the image and decrease the corresponding Gaussian kernel size, reducing the computational complexity of the processing. We recursively invoke Eq. \ref{eq3} to enable fast filtering. The image size is successively downsampled by half while simultaneously reducing the filter $\sigma_n$ size by half. The recursive process continues until the filter size reaches the designated minimum value $\sigma_{min}$, which is set to 10 in our method. The process of generating enhanced image $I_2$ can be formulated as: 
\begin{equation}
  \label{eq3}
  \begin{aligned}
  &G(x, y, \sigma_n)=\lambda \exp \left(-\left(x^2+y^2\right) / 2 \sigma_n^2\right), \\
  &I_2^i=\sum_{n=1}^N \omega_n R_n^i \\
  & =\sum_{n=1}^N \omega_n\left(\log \left(I^i(x, y)\right)-\log \left(G\left(x, y, \sigma_n\right) * I^i(x, y)\right)\right),
  \end{aligned}
  \end{equation}
where $N$ is the number of scales, we set $N = 3$ in this method. And $i \in\{R, G, B\}$ is the color channel of the RGB image, and $(x, y)$ is the pixel location on the image. Here, $\omega_n$ is the weight of each scale where $\omega_1 + \omega_2 +...+ \omega_n = 1$, and we set $\omega_1=\omega_2=...=\omega_n$ in this paper. The symbol $*$ denotes convolution,  and $G(x, y, \sigma_n)$ is the Gaussian filter. In this paper, we adopt three scales $\sigma_n \in [30, 150, 300], n \in \{1,2,3\}$. A smaller Gaussian kernel (e.g., $\sigma=30$) can better preserve image details and is suitable for detail enhancement and texture preservation. Larger Gaussian kernels (e.g., $\sigma=\{150,300\}$) can provide a wider smoothing effect and are suitable for noise removal. This ensures that the final image is a combination of the most important information from each scale. The efficient strategy allows for accelerated filtering while preserving the essential details of the image. The downsampling of the image and reduction in filter size ensure computational efficiency without compromising the overall filtering quality. The improved UIE model water-MSR can generate underwater restoration images in real-time, resulting in clearer and more natural underwater images.

\subsection{Cross-domain Feature Interaction Module}
\begin{figure*}[!t] 	
\centering 	
\includegraphics[width=\textwidth,height=\textheight,keepaspectratio]{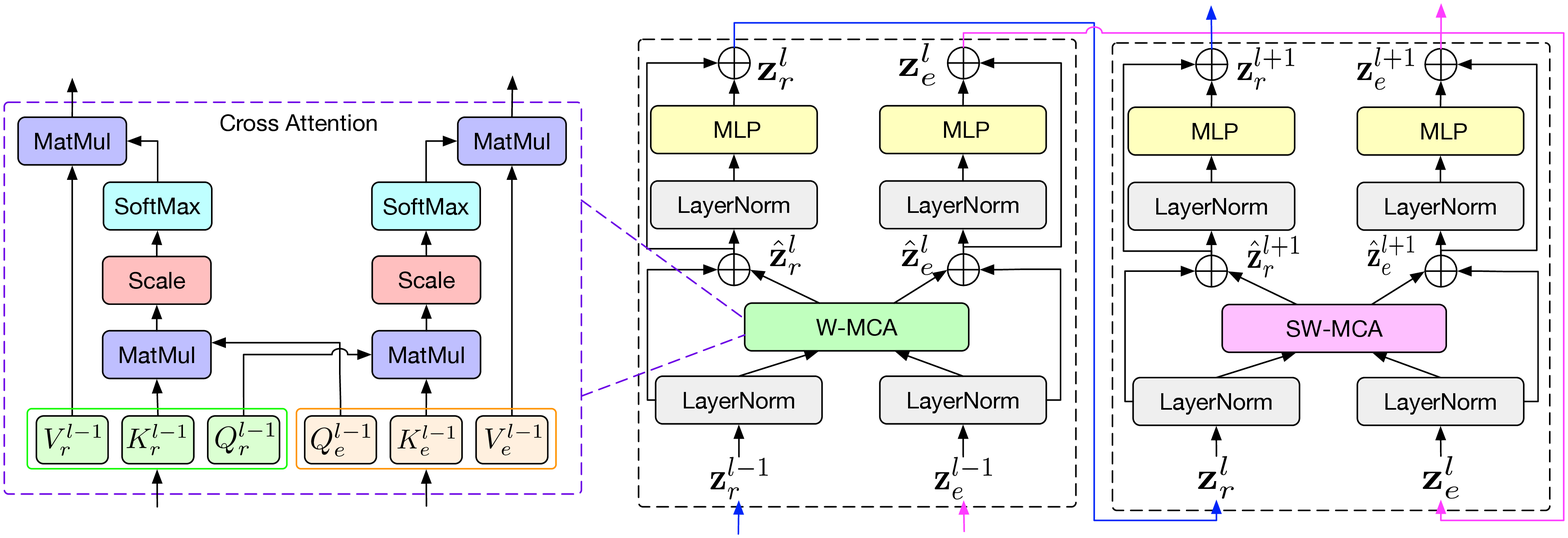}  	
\caption{Illustration of the proposed CFI module.} 	
\label{fig:cfi} 
\end{figure*}

Since the enhanced underwater image shows more appearance and discriminative information and can improve the visibility of small objects on the low-contrast areas, while the raw underwater image can preserve the natural characteristics of the scene and provide clear texture information about objects, it is important to interact the information of the two domains for UOD. In order to associate the two domains and leverage the complementary information, we design a cross-domain feature interaction (CFI) module based on the multi-head cross-attention (MCA), as shown in Fig. \ref{fig:cfi}. Starting from the raw image $I_1 \in \mathbb{R}^{H \times W \times C_{in}}$ and enhanced image $I_2 \in \mathbb{R}^{H \times W \times C_{i n}}$. $H$, $W$, and $C_{in}$ are the height, width, and channel number of input images. We first reshape the input into $\frac{HW}{M^2} \times M^2 \times C_{in}$ features by using a patch partition module. This module divides the input into non-overlapping $M \times M$ local windows, where $\frac{HW}{M^2}$ is the total number of windows. The window size is set to $M = 7$ by default. A linear embedding layer is applied to these features to project them to a desired dimension (denoted as $C$). Then, these patch tokens are sent into the proposed CFI module. To produce a hierarchical representation, we reduce the number of tokens by patch-merging layers as the network gets deeper. To enable the interaction between multi-domain information, we propose to use a multi-head cross-attention mechanism instead of a self-attention mechanism. As the computational complexity of MCA is quadratic to image size, this would be intractable on high-resolution images. Therefore, we build the CFI module by replacing the standard multi-head cross-attention module with a module based on shifted windows in Swin Transformer \cite{liu2021swin}. A CFI module consists of a shifted window-based MCA module, followed by a 2-layer MLP with GELU nonlinearity in between. A LayerNorm (LN) layer is applied before each MCA module and each MLP, and a residual connection is applied after each module. Unlike the original shifted window-based self-attention, the CFI module accepts two input vectors, enabling cross-domain global context exchange. 

Specifically, given feature $z_{r}$ of the raw image and the feature $z_e$ of the enhanced image, both with a size of $H \times W \times C$. For each local window feature $\{Z_{r}, Z_{e}\} \in \mathbb{R}^{M^2 \times C}$, we use three learnable weight matrices $\{W_r^Q, W_e^Q\} \in \mathbb{R}^{C \times C}$, $\{W_r^K, W_e^K\} \in \mathbb{R}^{C \times C}$, $\{W_r^V, W_e^V\} \in \mathbb{R}^{C \times C}$ that are shared across different windows to project it into the query $\{Q_r, Q_e\}$, key $\{K_r, K_e\}$ and value $\{V_r, V_e\}$ by:
\begin{equation}
\begin{aligned}
&\{Q_r, K_r, V_r\}=\left\{Z_r W_r^Q, Z_r W_r^K, Z_r W_r^V\right\}, \\
&\{Q_e, K_e, V_e\}=\left\{Z_e W_e^Q, Z_e W_e^K, Z_e W_e^V\right\}. \\
\end{aligned}
\end{equation}
Then, the cross-attention function essentially computes the dot product of the query with all keys and then normalizes the result using the softmax operator to generate attention scores. We follow \cite{liu2021swin} by including a relative position bias $B \in \mathbb{R}^{M^2 \times M^2}$ to each head in computing similarity. The cross-attention mechanism is defined as: 
\begin{equation} 
\begin{aligned}
&\operatorname{Attention}(Q, K, V)=\operatorname{SoftMax}\left(Q K^T / \sqrt{d}+B\right) V, \\
&\operatorname{MCA}(Z_r, Z_e) = (\operatorname{Attention}\left(Q_e, K_r, V_r\right), \\ &\operatorname{Attention}\left(Q_r, K_e, V_e\right)),
\end{aligned}
\end{equation} 
where $Q, K, V \in \mathbb{R}^{M^2 \times d}$ are the query, key and value matrices; $d$ is the $query/key$ dimension, and $M^2$ is the number of patches in a window. In our implementation, the process of multi-head cross-domain fusion unit is defined as: 
\begin{equation}
\begin{aligned}
\label{mca}
&\left(\tilde{a}_r^{l-1}, \tilde{b}_e^{l-1}\right)=\operatorname{W-MCA}\left(L N\left(z_r^{l-1}\right), L N\left(z_e^{l-1}\right)\right) \\
&\left(\tilde{a}_r^l, \tilde{b}_e^l\right)=\operatorname{SW-MCA}\left(LN\left(z_r^l\right), L N\left(z_e^l\right)\right), \\
\end{aligned}
\end{equation}
where $\tilde{a}_r^{l-1}$ and $\tilde{b}_r^{l-1}$ denote the output features of window-based MCA for block $l-1$ of raw image feature and enhanced image feature, respectively. And $\tilde{a}_r^l$ and $\tilde{b}_r^l$ denote the output features of shifted window-based MCA for block $l$. W-MCA and SW-MCA denote window-based multi-head cross-attention using regular and shifted window partitioning configurations \cite{liu2021swin}, respectively.
Based on the MCA, the whole process of the cross-domain feature interaction is defined as:
\begin{equation}
\begin{aligned}
\label{cfi}
&\hat{z}_r^l = \tilde{a}_r^{l-1} +z_r^{l-1}, \\
&\hat{z}_e^l=\tilde{b}_e^{l-1}+z_e^{l-1},\\
&z_r^l=\operatorname{MLP}\left(\operatorname{LN}\left(\hat{z}_r^l\right)\right)+\hat{z}_r^l,  \\ 
&z_e^l=\operatorname{MLP}\left(\operatorname{LN}\left(\hat{z}_e^l\right)\right)+\hat{z}_e^l,
% & \hat{\mathbf{z}}_r^{l+1}=\tilde{\mathbf{a}}_r^{l}+\mathbf{z}_r^l, \\ &\hat{\mathbf{z}}_e^{l+1}=\tilde{\mathbf{b}}_e^{l}+\mathbf{z}_e^l, \\
% & \mathbf{z}_r^{l+1}=\operatorname{MLP}\left(\mathrm{LN}\left(\hat{\mathbf{z}}_r^{l+1}\right)\right)+\hat{\mathbf{z}}_r^{l+1}, \\
% & \mathbf{z}_e^{l+1}=\operatorname{MLP}\left(\mathrm{LN}\left(\hat{\mathbf{z}}_e^{l+1}\right)\right)+\hat{\mathbf{z}}_e^{l+1},
\end{aligned}
\end{equation}
where $\hat{z}_r^l$ and $z_r^l$ denote the output features of the residual block and the MLP module for block $l$ of a raw image feature, respectively. While $\hat{z}_e^l$ and $z_e^l$ denote the output features of the residual block and the MLP module for block $l$ of an enhanced image feature.

As presented in Eq. \ref{mca} and \ref{cfi}, for $Q_e$ from the enhanced image, it incorporates cross-domain information by performing attention weighting with $K_r$ and $V_r$ from the raw domain, while preserving information in the enhanced image through the residual connection and vice versa. Our model can integrate global cross-domain interactions.

\subsection{Gated Feature Fusion Module}
In the previous section, we extend the Swin-Transformer encoding backbone with the CFI module to exchange information globally between the two domains. In this fusion stage, we merge the features from two domains into a single feature map. Taking into account that there may exist contaminations from unreliable generated results of water-MSR, directly integrating the cross-domain information may induce negative results. To cope with these issues, we design a gated feature fusion (GFF) module to control the fusion rate of cross-domain information by using a gate function. Specially, the output interaction feature map $z_r$ and $z_e$ is associated with a gated map $G_r \in[0,1]^{H \times W \times C}$ and $G_e \in[0,1]^{H \times W \times C}$, respectively. With these gated maps, the mechanism of the GFF module can be expressed with the following equations:
\begin{equation}
\begin{aligned}
\label{gff}
&F^s= \left(G_r^s \odot z_r^s\right) \oplus \left(G_e^s \odot z_e^s\right), \\
&G_r^s = sigmoid(w_r^s \cdot z_r^s), \\
&G_e^s = sigmoid(w_e^s \cdot z_e^s),
\end{aligned}
\end{equation}
where $F^s$ is the output feature of the $s^{th}$ stage of GFF module. Here, $s \in \{1,2,3,4\}$, $\oplus$ represents the operation of feature addition, and $\odot$ represents the feature multiplication. The $w_r^s$ and $w_e^s$ are parameterized by a convolutional layer, respectively. 

With the learned gated function, we can effectively aggregate the cross-domain advantageous information of the raw and enhanced images, and the gate function controller focuses on regulating the fusion rate of the cross-domain information. As shown in Fig. \ref{fig:framework}, the number of GFF modules is 4. The output resolutions of the four stages are $\left(\frac{H}{4} \times \frac{W}{4}\right)$, $\left(\frac{H}{8} \times \frac{W}{8}\right)$, $\left(\frac{H}{16} \times \frac{W}{16}\right)$, and $\left(\frac{H}{32} \times \frac{W}{32}\right)$, respectively. Here, $H, W$ are the height and width of input images. These stages jointly produce a hierarchical representation, and can conveniently leverage different existing detection heads for detection tasks.

\subsection{Detection Head}
Our method can be applied to many existing popular two-stage or single-stage detectors by combining with the different detection heads. After getting the hierarchical feature maps, they will be sent to a detection head to output the localization and classification. In this paper, we adopt the detection head of AutoAssign \cite{zhu2020autoassign}, as shown in Fig. \ref{fig:framework}. The detection head of AutoAssign can achieve appearance-aware through a fully differentiable weighting mechanism. And the confidence weighting of Autoassign can be used to adjust the specific assign strategy of each object instance. This detection head can adapt well to various underwater objects of varying sizes and shapes in the UOD task. In our experiments, we also validate the performance of combining the proposed method with other detection heads (e.g., Faster R-CNN \cite{FasterRCNN} and Deformable DETR \cite{deformable}), and achieve good performance as well. The proposed framework is optimized in an end-to-end manner. Different detection heads are equipped with different loss functions. For example, when we use the Autoassign detection head, the loss function is the same as the loss function in AutoAssign.

\section{Experiments}
\subsection{Datasets}
We conduct experiments on four challenging underwater datasets (DUO \cite{DUO}, Brackish \cite{brackish}, Trashcan \cite{hong2020trashcan}, and WPBB \cite{EfficientDets}) to validate the performance of our method.

\textbf{DUO} \cite{DUO} is a newly published underwater dataset that contains a variety of underwater scenes and more reasonable annotations. It is a refined version of the UTDAC2020 dataset \cite{chen2022swipenet}, which is from Underwater Target Detection Algorithm Competition 2020. It contains $7,782$ images ($6,671$ images for training; $1,111$ for testing), and $74,515$ instances among $4$ common categories: echinus, holothurian, starfish, and scallop. The images in the dataset contain four resolutions: $3840 \times 2160, 1920 \times 1080, 720 \times 405$, and $586 \times 480$. 

\textbf{Brackish} \cite{brackish} is the first annotated underwater image dataset captured in temperate brackish waters. It contains $6$ classes: big fish, crab, jellyfish, shrimp, small fish, and starfish. The training set, validation set, and test set are randomly split into $9,967$, $1,467$, and $1,468$ images, respectively. The image size is $960 \times 540$.

\textbf{TrashCan} \cite{hong2020trashcan} is the first instance-segmentation annotated dataset of underwater trash. It contains 16 classes, including trash, ROVs, and a wide variety of undersea flora and fauna, etc. The training set and test set are randomly split into $6008$ and $1204$, respectively.

\textbf{WPBB} \cite{EfficientDets} is a detection dataset of in-water plastic bags and bottles, which is comprised of annotated images ($900$ images currently). It contains $2$ common categories: plastic bags and plastic bottles. The training set and test set are randomly split into $720$ and $180$, respectively.

\subsection{Experimental Details and Evaluation Metrics}
\textbf{Experimental Details:} Our method is implemented on MMdetection \cite{mmdetection}. For our main results, we use multi-scale training with the long edge set to $1300$ and the short edge set to $800$. The training schedule is 36 epochs with an initial learning rate of $2.5 \times 10^{-3}$ and the learning rate is decreased by 0.1 after 27 and 33 epochs, respectively. Our method is trained on 2 GeForce RTX 3090 GPUs with a total batch size of 4 (2 images per GPU) for training and a single 3090 GPU for inference. We adopt the AdamW \cite{loshchilov2017decoupled} for training optimization, where the weight decay is 0.0001 and the momentum is 0.9.  In the experiments, no data augmentation except the traditional horizontal flipping is utilized. All other hyper-parameters follow the settings in MMdetection.

\textbf{Evaluation Metrics:} The main reported results in this paper follow standard COCO-style Average Precision (AP) metrics that include $AP_{50}$ (IoU = 0.5), $AP_{75}$ (IoU = 0.75) and AP. AP is measured by averaging over multiple IoU thresholds, ranging from 0.5 to 0.95 with an interval of 0.05. 
\subsection{State-of-the-art Comparison}
We compare GCC-Net against some state-of-the-art methods on the four underwater datasets. The results are shown in Table \ref{tab1} and Table \ref{tab2}.
\subsubsection{Results on DUO}
We evaluate the effectiveness of GCC-Net with comparisons against the baseline method \cite{zhu2020autoassign} and the state-of-the-art methods. For a fair comparison, we divide the compared methods into two groups, i.e., the generic object detectors and the underwater object detectors. Table \ref{tab1} summarizes the results of the DUO dataset, from which we draw two observations.  
\begin{table*}[!t]
\centering
\caption{Comparisons with state-of-the-art on the DUO dataset. The results with red and blue colors indicate the best and second-best results of each column, respectively.}
\label{tab1}
\begin{adjustbox}{max width=\textwidth}   
\begin{tblr}{
	vline{2,5,9} = {-}{},
	hline{1-2,11,18} = {-}{},
}
Methods  & AP    & $AP_{50}$ & $AP_{75}$ & echinus & starfish & holothurian & scallop & FPS \\
\textbf{\textit{Generic Object Detector:}}    &  &        &        &         &          &             &         &     \\
Faster R-CNN \cite{FasterRCNN}   & 61.3 & 81.9  & 69.5  & 70.4 & 71.4 & 61.4& 41.9 &19.2 \\
Cascade R-CNN \cite{CascadeRCNN}   &61.2 &  82.1 & 69.2& 69.0 &  72.0 &61.9  & 41.9  & 17.2\\
AutoAssign \cite{zhu2020autoassign}  & 66.1 & 85.7  & 72.6  & 74.1  & 75.5    & 65.8       & 48.9    & 18.5     \\
SABL w/ Cascade R-CNN \cite{sabl} & 63.4 & 81.2  & 70.5  & 72.0   & 74.0    & 64.7& 42.8   & 9.6    \\
DetectoRS \cite{qiao2021detectors} & 64.8 & 83.5  & 72.4  & 73.5   & 74.3    & 65.8  & 45.7    &  6.9   \\
Deformable DETR \cite{deformable}   & 63.7 & 84.4  & 71.9  & 71.6   & 73.9    & 63.0  & 46.3  &  16.0   \\
GFL \cite{gfl} & 65.5 & 83.7  & 71.9  & \textcolor{blue}{74.2}   & \textcolor{blue}{75.9}    & 64.3       & 47.5   &19.8 \\
YOLOv7 \cite{wang2022yolov7} & \textcolor{blue}{66.3} & \textcolor{blue}{85.8}  & \textcolor{blue}{73.9}  & 73.7   & 74.5     & 66.3        & \textcolor{blue}{50.8}    & \textcolor{red}{30.1}    \\
\textbf{\textit{Underwater Object Detector:}} & & &  & & & &  &     \\
ROIMIX \cite{ROIMIX}      & 61.9 & 81.3  & 69.9  & 70.7   & 72.4    & 63.0       & 41.7   &  19.2    \\
ERL-Net \cite{dai2023edgeguided}     & 64.9 & 82.4  & 73.2  & 71.0   & 74.8    & \textcolor{blue}{67.2}       & 46.5   &  8.7   \\ 
Boosting R-CNN \cite{song2023boosting}   & 63.5 & 78.5  & 71.1  & 69.0   & 74.5    & 63.8 & 46.8    & \textcolor{blue}{22.3}    \\
SWIPENet \cite{chen2022swipenet}  & 63.0 & 79.7  & 72.5  & 68.5   & 73.6    & 64.0       & 45.9   &  4.1    \\
RoIAttn \cite{liang2022excavating}  & 62.3 & 82.8  & 71.4  & 70.6   & 72.6    & 63.4       & 42.5   &  9.8   \\
GCC-Net (Ours)  & \textcolor{red}{69.1} & \textcolor{red}{87.8}  & \textcolor{red}{76.3}  & \textcolor{red}{75.2}   & \textcolor{red}{76.7}    & \textcolor{red}{68.2}       & \textcolor{red}{56.3}   & 15.6     
\end{tblr}
\end{adjustbox}
\end{table*}
First, for the accuracy measured by AP, we achieve 69.1\% AP. Comparing GCC-Net against the baseline method AutoAssign \cite{zhu2020autoassign}, the proposed method outperforms AutoAssign by 3.0\% (66.1\% vs 69.1\%). By using with the same backbone Swin-T \cite{liu2021swin}, GCC-Net achieves the best result among GOD methods (e.g., DetectoRS \cite{qiao2021detectors} and YOLOv7 \cite{wang2022yolov7}) and UOD methods (e.g., SWIPENET \cite{chen2022swipenet} and Boosting R-CNN \cite{song2023boosting}). Specifically, GCC-Net outperforms DetectoRS by 4.3 \% (64.8\% vs 69.1\%), Deformable DETR by 5.4\% (63.7\% vs 69.1\%), GFL by 3.6 \% (65.5\% by 69.1\%), and YOLOv7 by 2.8\% (66.3\% vs 69.1\%), which is a large margin. 

Regarding the comparison with UOD methods, we select the recent open-source methods to compare with our method. The proposed method also achieves the best result among these methods. Specifically, GCC-Net surpasses SWIPENET by 6.1 \% (63.0\% vs 69.1\%), Boosting R-CNN by 5.6\% (63.5\% vs 69.1\%), RoIAttn by 6.8\% (62.3\% vs 69.1\%), and ERL-Net by 4.2\% (64.9\% vs 69.1\%). The experimental results demonstrate the superior performance of our method in underwater object detection. Complicated underwater environments bring new challenges to object detection, such as low contrast and unbalanced light conditions. Our method effectively addresses these issues, making the objects in low-contrast regions clearly visible and thereby improving the performance of UOD tasks.

Second, under the same setting, we compare the inference speed of different methods on the DUO dataset. The hardware platform for testing is a GeForce RTX 3090 with a batch size of 1. We adopt single-scale testing in the experiment. As shown in Table \ref{tab1}, we achieve 15.6 FPS. Our method demonstrates comparable processing speed to other methods, fulfilling the real-time operational requirement.

\begin{table*}
\centering
\caption{Benchmarking results between GCC-Net and other state-of-the-art methods on Brackish, TrashCan, and WPBB datasets. The results with red and blue colors indicate the best and second-best results of each column, respectively.}
\label{tab2}
\begin{adjustbox}{max width=0.8\textwidth}  
\begin{tblr}{
row{2} = {c},
cell{1}{1} = {r=2}{},
cell{1}{2} = {c=2}{c},
cell{1}{4} = {c=2}{c},
cell{1}{6} = {c=2}{c},
vline{2,4} = {1-2}{},
vline{4,6} = {1-2}{},
vline{2,4,6} = {3-11}{},
hline{1,11} = {-}{},
hline{1-3} = {1-7}{},
}
Methods & Brackish &        & TrashCan &        & WPBB  &        \\
   & AP       & $AP_{50}$ & AP       & $AP_{50}$ & AP    & $AP_{50}$ \\
Faster R-CNN \cite{FasterRCNN}         &  61.2    & 62.9  & 31.2 & 55.3 & 75.7 & 98.7  \\
SABL w/ Cascade R-CNN \cite{sabl} &  80.4    & \textcolor{blue}{98.7}  & 34.1    & 54.3  & \textcolor{blue}{80.2} & \textcolor{blue}{99.4}  \\
YOLOv7  \cite{wang2022yolov7} &  57.2  & 88.5    & 24.1      & 43.4    & 78.7 & \textcolor{red}{99.5}  \\
Deformable DETR \cite{deformable} &77.5& 97.1  & 36.1      & 56.9   & 73.2 & 98.3   \\
Boosting R-CNN \cite{song2023boosting}& 79.6    & 97.4  &   36.8       &  57.6      & 78.5 & 97.1  \\
RoIAttn \cite{liang2022excavating}   & 78.3    & 91.0  &  32.6  & 57.2  & 70.2 & 88.1  \\
ERL-Net \cite{dai2023edgeguided}  & \textcolor{red}{85.4}    & \textcolor{red}{98.8}  & \textcolor{blue}{37.0}    & \textcolor{blue}{58.9}  & 79.7 & 98.5  \\
GCC-Net & \textcolor{blue}{80.5}    & 98.3  & \textcolor{red}{41.3} & \textcolor{red}{61.2}    & \textcolor{red}{81.0} & \textcolor{red}{99.5}
\end{tblr}
\end{adjustbox}
\end{table*}

\subsubsection{Results on TrashCan and WPBB}
These two datasets both include underwater debris of various types. The experimental results are presented in Table \ref{tab2}. The results demonstrate outstanding performance on both datasets, surpassing Faster R-CNN \cite{FasterRCNN}, SABL \cite{sabl}, YOLOv7 \cite{wang2022yolov7}, Deformable DETR \cite{deformable}, Boosting R-CNN \cite{song2023boosting}, RoIAttn \cite{liang2022excavating}, and et al. Specifically, GCC-Net achieves 41.3\% AP on the TrashCan dataset, surpassing the state-of-the-art method ERL-Net by 4.3\%. And GCC-Net achieves 81.0\% AP on the WPBB dataset, surpassing the state-of-the-art method SABL by 0.8\%. The aquatic environments in the TrashCan dataset are predominantly deep blue, with a higher concentration of suspended particles resulting in relatively blurred images. In contrast, the WPBB dataset includes underwater images captured in shallow blue waters. The experimental results demonstrate that our method achieves outstanding detection performance in various underwater environments without the need to retrain the UIE model, thus highlighting the efficacy and robustness of the proposed method. 
\subsubsection{Results on Brackish}
The comparison results on Brackish are shown in Table \ref{tab2}. GCC-Net achieves 80.5\% AP and 98.3\% $AP_{50}$, outperforming most GOD and UOD methods. Note that our method does not surpass ERL-net \cite{dai2023edgeguided} on the Brackish dataset, we believe that this is due to the fact that the Brackish dataset \cite{brackish} was collected in a brackish water environment, which is characterized by a gray or blackish appearance, rather than the typical blue or green colors commonly presented in underwater environments. Consequently, the UIE model may have exhibited reduced efficacy in this environment. The ERL-Net achieves improved performance on the brackish dataset by employing a multi-level feature approach, but it sacrificed detection efficiency. In contrast, our proposed method achieves higher inference speed compared to ERL-Net, which is a critical factor for the practical deployment of underwater robotic platforms. This advantage makes our method more suitable for real-world applications.

\subsection{Ablation Studies}
In this section, we conduct a series of ablation experiments on the DUO dataset to evaluate the effectiveness of our proposed method. We choose AutoAssign \cite{zhu2020autoassign} as the baseline for ablations due to its impressive performance and efficient inference speed.
\begin{table}[t]
\centering
\caption{Ablation studies of only using raw images or enhanced images. The DUO dataset is used in this experiment. The bold results indicate the best performance.}
\label{tab3}
\begin{adjustbox}{max width=0.45\textwidth}   
\begin{tblr}{
column{even} = {c},
column{1} = {c},
column{3} = {c},
column{5} = {c},
vline{2,6-7} = {-}{},
hline{1-2,6} = {-}{},
} Domain & echinus & starfish & holothurian & scallop & AP   & Time(s) \\
Raw &  \textbf{74.1}  & \textbf{75.5}    & \textbf{65.8}       & \textbf{48.9}   & \textbf{66.1} &  - \\
MMLE \cite{zhang2022underwater}             & 69.3   & 72.9    & 59.1       & 44.8   & 64.3 & 1.72s    \\
FUnIE-GAN \cite{islam2020fast}           & 46.2   & 52.4    & 37.8       & 48.4   & 46.2 & 0.10s    \\
% IMSRCP \cite{tang2019efficient} & -   & -    & -       & -   & - & 20s\\
water-MSR (ours) & 73.6   & 73.6    & 62.9       & 48.4   & 64.6 & 0.12s   
\end{tblr}
\end{adjustbox}
\end{table}
\begin{table}[t]
  \centering
  \caption{Ablation studies of using different image domains. The DUO dataset is used in this experiment. The bold results indicate the best performance.}
  \label{uie}
  \begin{adjustbox}{max width=0.45\textwidth}   
  \begin{tblr}{
    vline{3,7} = {-}{},
    hline{1-2,6} = {-}{},
  }
  Domain 1 & Domain 2         & echinus & starfish & holothurian & scallop & AP  \\
  Raw      & Raw              & 74.0  & 74.3 &65.0   &54.7 &67.0 \\
  Raw      & Color Jittering  & 73.3   &  73.5    &  63.6      &  50.4   & 65.2   \\
  % Raw      & FUnIE-GAN \cite{islam2020fast}        &   \\
  Raw      & MMLE \cite{zhang2022underwater}   & 74.2& 75.6 & 67.1 & 54.5& 67.8 \\
  Raw      & water-MSR (ours) & \textbf{75.2} & \textbf{76.5} & \textbf{68.1} & \textbf{56.4} & \textbf{69.1}
  \end{tblr}
  \end{adjustbox}
  \end{table}

\subsubsection{Effect of Underwater Image Enhancement (UIE) Model} 
To demonstrate the role of the UIE method in the UOD task, we conduct two experiments: (1) exploring the performance of only using the enhanced images as input of UOD, as shown in Table \ref{tab3}; (2) using images generated by different UIE or color jittering methods as another image domain. This setting aims to validate that the performance improvement lies in the interaction and collaboration of the raw and enhanced images rather than resulting from increased input images and network parameters, as shown in Table \ref{uie}.
\begin{table}[t]
  \centering
  \caption{Ablation studies of proposed modules in GCC-Net. The DUO dataset is used in this experiment. The bold results indicate the best performance.}
  \label{modules}
  \begin{adjustbox}{max width=0.45\textwidth} 
  \begin{tblr}{
    column{2} = {c},
    column{3} = {c},
    column{4} = {c},
    column{5} = {c},
    vline{2,6} = {-}{},
    hline{1-2,8} = {-}{},
  }  Method & MFA & CFI & FF & GFF & AP & $AP_{50}$ \\
  AutoAssign &   &   &  & & 66.7    & 85.8     \\
  1    & \cmark & & &   &  65.1  & 84.5     \\
  2   & \cmark & & \cmark & &  66.9   & 86.0   \\
  3   & & \cmark &   &  &  68.1   &  86.9    \\
  4    &  & \cmark & \cmark & & 68.6  &  87.2   \\
  GCC-Net (ours)    & & \cmark &  & \cmark & \textbf{69.1}& \textbf{87.8}    
  \end{tblr}
  \end{adjustbox}
  \end{table}

For the first ablation experiment, we use three UIE methods (MMLE \cite{zhang2022underwater}, FUnIE-GAN \cite{islam2020fast}, and the proposed water-MSR) to generate enhanced images and evaluate the object detection performance. AutoAssign \cite{zhu2020autoassign} is used in this experiment. MMLE is a physical UIE model that adjusts the color and details of an input image according to a minimum color loss principle and a maximum attenuation map-guided fusion strategy. The code of MMLE \cite{zhang2022underwater} is encrypted and implemented in Matlab, so we run the model offline. FUnIE-GAN \cite{islam2020fast} is a real-time deep learning-based UIE method, we use its pre-trained model due to the lack of paired distorted and clear image sets. As shown in Table \ref{tab3}, the performance of only using the enhanced images as input is inferior to directly using the raw images. This result is attributed to the possibility of enhanced images losing critical patterns or introducing noise that hampers the detection process. Moreover, the processing time of MMLE is approximately 1.72 seconds per image, which does not meet the requirement for real-time performance. In addition, we compare the speed between the baseline UIE model IMSRCP and the proposed water-MSR model. When we apply the IMSRCP method with a Gaussian kernel size of 300 to an image of size $1300\times750$, the processing time is 20 seconds. However, with our improved water-MSR method, the processing time for a single image decreases to 0.12 seconds. This demonstrates that the recursive fast filtering strategy significantly reduces the model's processing speed.

\begin{figure*}[!t] 	
  \centering 	
  \includegraphics[width=\textwidth,height=\textheight,keepaspectratio]{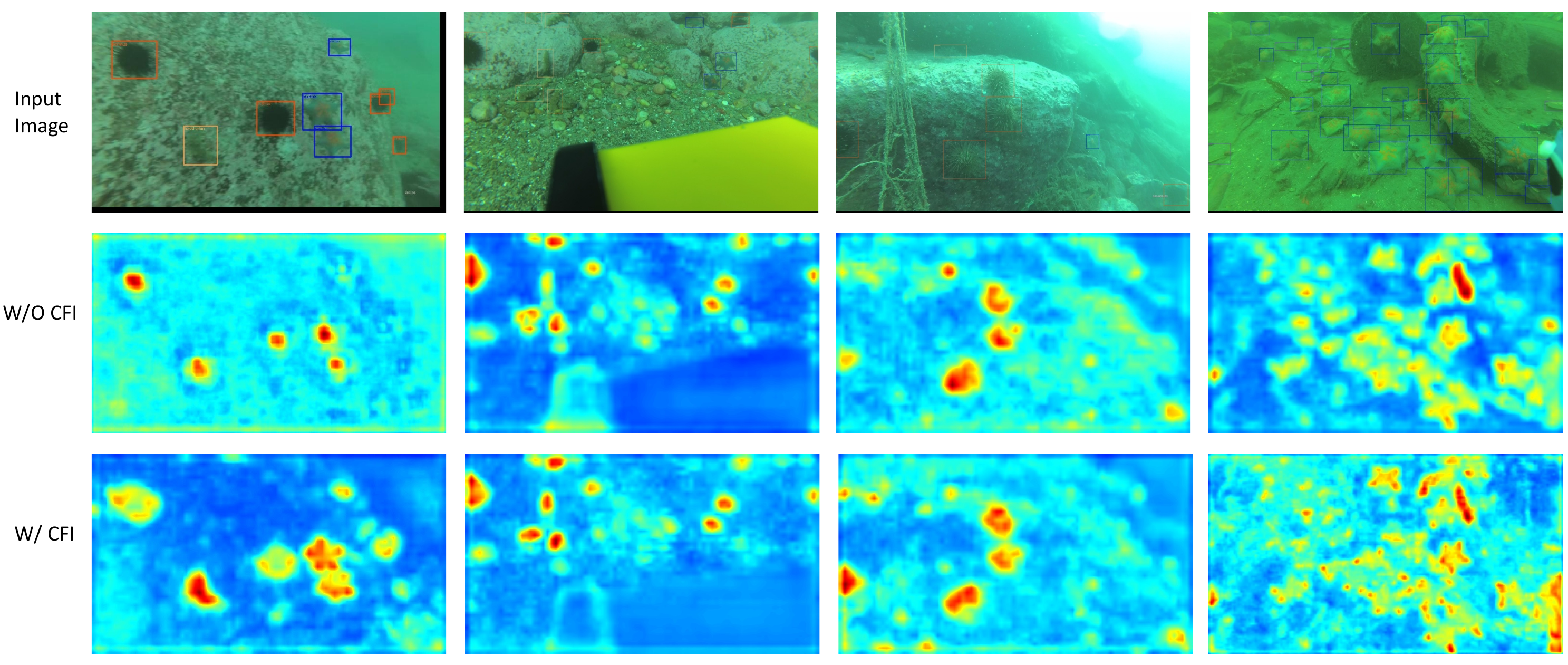}  	
  \caption{Visualization comparison results of feature map without (the second row) and with (the third row) CFI module by using Grad-CAM \cite{selvaraju2017grad}. Best viewed in color and with zoom.} 	
\label{fig:cam} 
\end{figure*}

For the second ablation experiment, we drive four settings, as shown in Table \ref{uie}. In the first configuration, we input the raw image twice to investigate whether the performance difference is caused by the difference in network structure. The accuracy under this configuration is 67.0\%, similar to the results obtained by inputting only one raw image (66.1\%). The result indicates that the performance improvement of our method does not simply rely on changing the network structure and parameters. In the second configuration, we use random color jittering to randomly change the brightness, color, contrast, and saturation of the raw image, the AP of this configuration is 65.2\%. This result demonstrates that our method does not rely on increasing input data to achieve performance improvement. Next, we compare the performance of using different enhanced images as another image domain, generated by water-MSR and other UIE methods (e.g., MMLE \cite{zhang2022underwater}). The AP is 69.1\% and 67.8\%, respectively. The results demonstrate the advantage of the interaction and fusion between the raw images and the enhanced images for underwater object detection.
\begin{table}[t]
  \centering
  \caption{Ablation Studies of proposed method GCC-Net combined with different detection heads. The DUO dataset is used in this experiment.}
  \label{head}
  \begin{adjustbox}{max width=0.45\textwidth} 
  \begin{tblr}{
    hline{1-2,5,9} = {-}{},
  }
  Methods                    & AP & $AP_{50}$ & Params (M) & FLOPs (G) \\
  Faster R-CNN               & 61.3   & 81.9  & 44.76& 210.32   \\
  Deformable DETR            & 63.7 & 84.4   & 40.59   & 200.71\\
  AutoAssign                 & 66.1  & 85.7  & 36.74 & 208.18    \\
  Ours                       &    &         &   &  \\
  GCC-Net w/ Faster R-CNN    & 65.6 (\textcolor{blue}{+4.3})   &   86.6 (\textcolor{blue}{+4.7}) & 44.76   & 302.94 \\
  GCC-Net w/ Deformable DETR & 66.9 (\textcolor{blue}{+3.2})   &  86.9 (\textcolor{blue}{+2.5})    &  40.59   & 293.33\\
  GCC-Net w/ AutoAssign      & 69.1 (\textcolor{blue}{+3.0})  &   87.8 (\textcolor{blue}{+2.1})  & 36.74 &  300.79
  \end{tblr}
  \end{adjustbox}
  \end{table}

\subsubsection{Effect of Cross-domain Feature Interaction (CFI) Module} 
To investigate the importance of the CFI module, we perform a study of different interaction strategies. We compare the performance of two feature interaction strategies: multi-layer feature addition (MFA) and the proposed CFI module. For the MFA strategy, features from different image domains are directly added together at each transformer stage. Taking the Swin-T model as an example, the implementation of the MFA strategy can be summarized as follows: one backbone network is used to extract features of the raw images, while the other backbone network is used to extract features of the enhanced images. Then, we perform element-wise addition on the features at each layer to obtain the fused features. As shown in Table \ref{modules}, the performance of using the MFA strategy is 65.1\%, which is lower than using the proposed CFI module. The performance of using the CFI module is 68.1\%, resulting in a performance gain of 3.0\%. Furthermore, the MFA strategy aims to leverage two independent backbone networks to generate multi-domain features and fuse them together at each layer. Compared to the proposed CFI module, the MFA strategy involves a significant increase in the number of parameters, as each backbone network has its own set of parameters. In contrast, we employ a unified module to interact with features from different domains, maintaining the same number of parameters as AutoAssign while achieving superior performance.

For a better understanding of the role of the proposed modules, we visualize the feature map of with/without CFI module in the backbone network by using Grad-CAM \cite{selvaraju2017grad}, as shown in Fig. \ref{fig:cam}. These attention maps are computed based on the activations of each spatial unit in the second stage of the backbone network, effectively indicating where the network focuses its attention to classify and localize objects within the input image. It is evident that the proposed module exhibits a heightened emphasis on object regions, particularly in areas of low contrast. Our method can accurately identify objects even under challenging conditions.
\begin{figure*}[!t] 	
  \centering 	
  \includegraphics[width=\textwidth,height=\textheight,keepaspectratio]{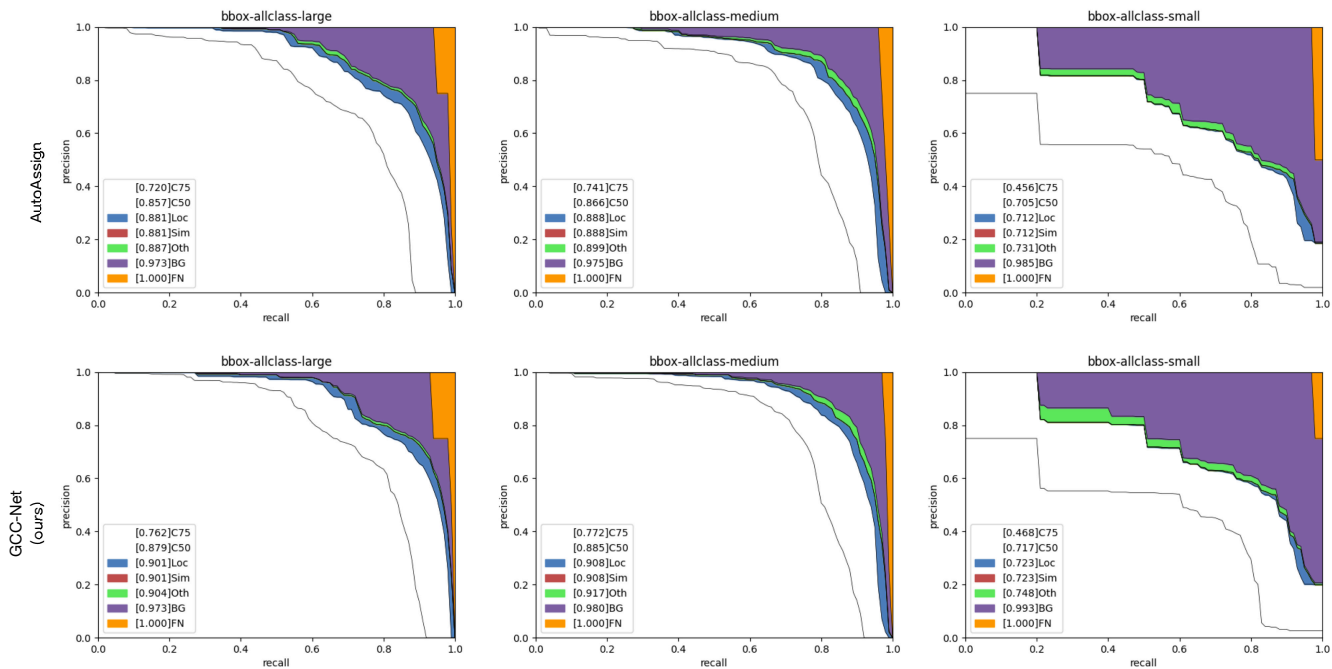}  	
  \caption{Error analysis plots of the baseline method AutoAssign \cite{zhu2020autoassign} (top row) and our method GCC-Net (bottom row) across three categories, on the large-sized (first column), the medium-sized objects (second column), and small-sized objects (the last column). As defined in \cite{lin2014microsoft}, a series of precision-recall curves with different evaluation settings are shown in each sub-image plot.} 	
  \label{fig:coco_error} 
  \end{figure*}

\subsubsection{Effect of Gated Feature Fusion (GFF) Module} 
To validate the effectiveness of the GFF module, we derive two settings: with the feature fusion (FF) module and with the GFF module. 
For the FF module, we add the outputs of the CFI module without considering adaptive fusion by setting both $G_r^s$ and $G_e^s$ in Eq. \ref{gff} to 1. As shown in Table \ref{modules}, the performance of using the FF module is 68.6\%, while using the GFF module is 69.1\%, resulting in a performance gain of 0.5\%. The results indicate that compared to directly merging the information from both domains, the proposed gated fusion strategy is more effective. The proposed module allows for the control of the fusion rate for each domain feature, mitigating the contamination caused by unreliable information. 

\subsubsection{Effect of Different Detection Head} 
Our method can be easily extended to various object detection frameworks. In this experiment, we apply the proposed method to different detection heads, including anchor-free (AutoAssign \cite{zhu2020autoassign}), anchor-based (Faster R-CNN \cite{FasterRCNN}), and transformer-based (Deformable DETR \cite{deformable}) methods, to validate its performance. As shown in Table \ref{head}, the proposed GCC-Net combined with Faster R-CNN surpasses the original Faster R-CNN by 4.3 points (65.6\% vs 61.3\%), the proposed GCC-Net combined with Deformable DETR surpasses the original Deformable DETR by 3.2 points (66.9\% vs 63.7\%), and the proposed GCC-Net combined with AutoAssign surpasses the original AutoAssign by 3.0 points (69.1\% vs 66.1\%). Moreover, we also compare the differences in parameter count and FLOPs (floating-point operations) between the proposed GCC-Net and the baselines. Parameters represent the number of learnable parameters in the model, including weights, biases, and other trainable parameters. FLOPs serve as a metric to measure computational load, indicating the number of floating-point operations required during program execution. Higher FLOPs values typically indicate a heavier computational workload. From Table \ref{head}, we observe that the proposed method exhibits a higher computational load, as it achieves an increased number of FLOPs compared to the baselines. It is important to emphasize that despite the increase in computational complexity, the parameter count remains unchanged. The difference in FLOPs suggests that our method requires more computational resources during the training process. However, it is crucial to note that the parameter count, which directly affects memory consumption, remains consistent. Moreover, our method consistently outperforms all baseline methods with a nearly 5 points improvement, demonstrating a significant performance enhancement. The results demonstrate the robust scalability of our method, positioning it as a potential new benchmark in the field. With its plug-and-play nature, our method can serve as a readily applicable module for future works to build upon.

With these ablation studies, we conclude that in the GCC-Net design: the online water-MSR model, cross-domain feature interaction, and gated feature fusion module all play essential roles in the final performance.
\begin{figure*}[] 	 
  \centering 	 \includegraphics[width=0.95\textwidth,height=0.95\textheight,keepaspectratio]{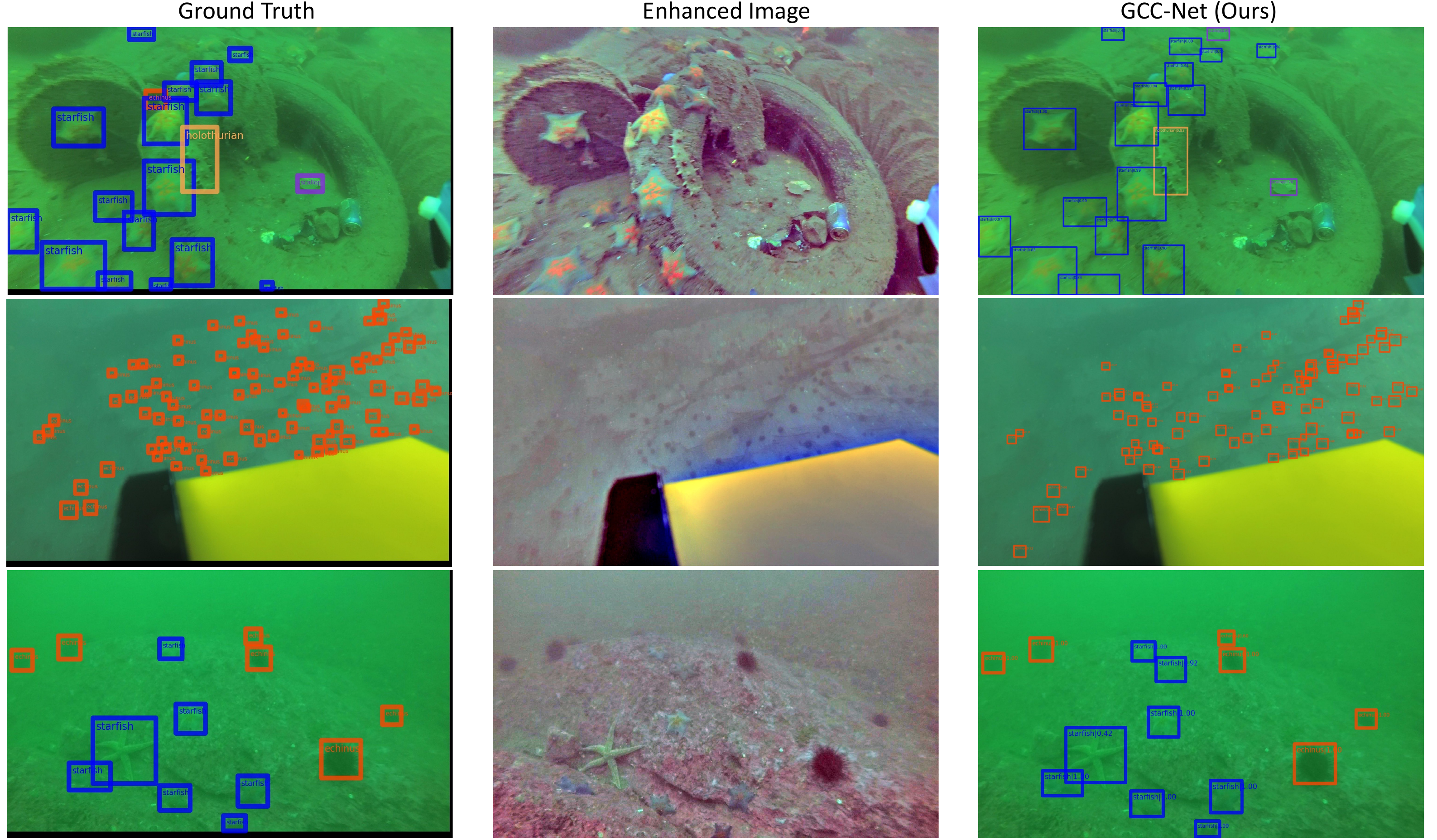} 
  \caption{The qualitative comparison results on the DUO dataset\cite{DUO}. The enhanced images are generated by water-MSR. The confidence threshold is set to 0.3 when visualizing these results. One color stands for one object class. Best viewed in color and with zoom.} 	 
\label{fig:duo}  
\end{figure*}
\begin{figure*}[!t] 	
  \centering 	
  \includegraphics[width=0.9\textwidth,height=0.9\textheight,keepaspectratio]{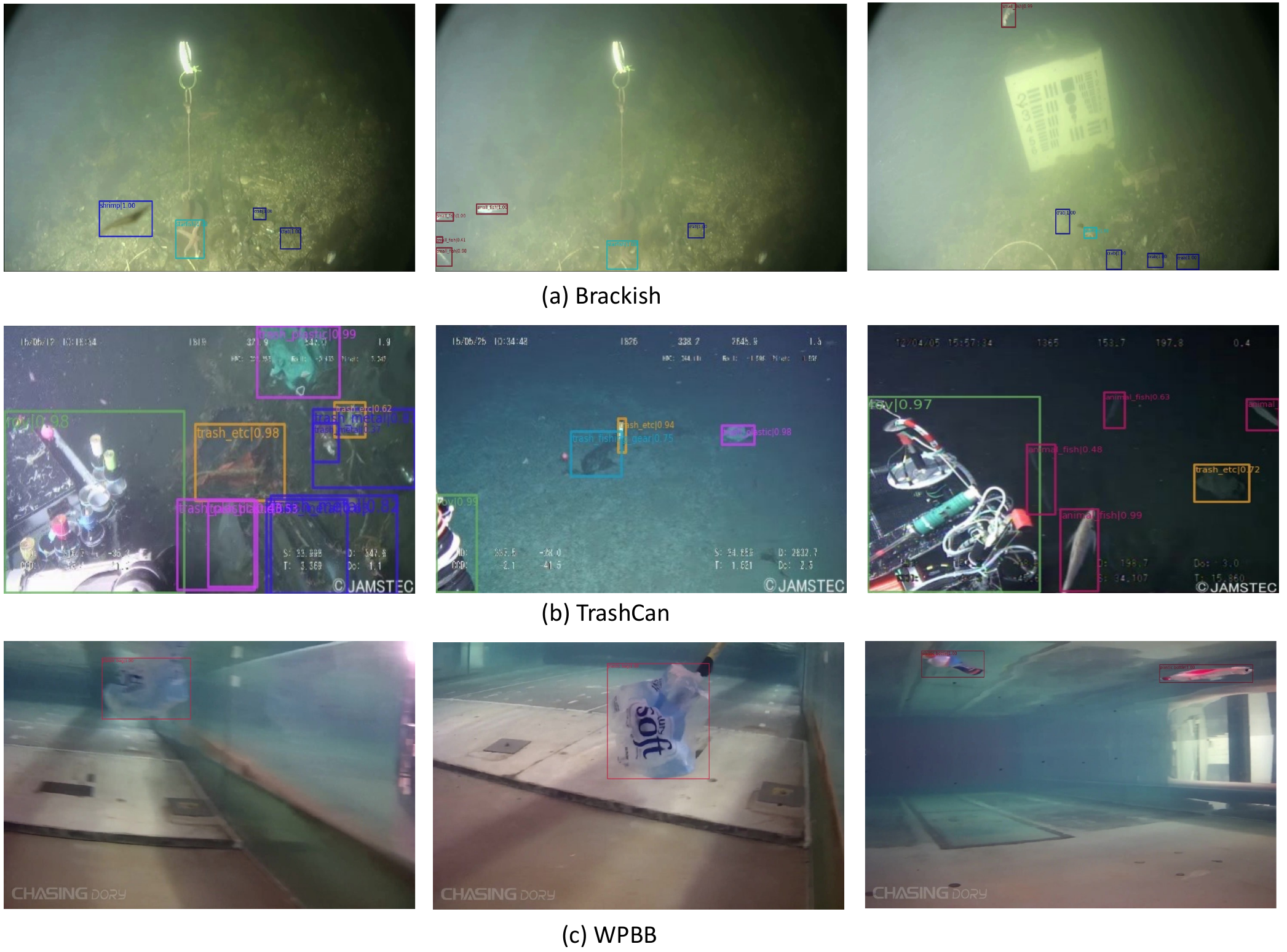}  	
  \caption{Detection results on Brackish\cite{brackish} (the first row), TrashCan \cite{hong2020trashcan} (the second row), and WPBB \cite{EfficientDets} (the third row). Best viewed in color and with zoom.} 	
  \label{fig:others} 
  \end{figure*}
\subsection{Error Analysis}
To further analyze the error types, we utilize the error analysis tool provided by mmdetection \cite{mmdetection}. Fig. \ref{fig:coco_error} shows the error plots on the DUO dataset. The plots in each sub-image represent a series of precision-recall curves with various evaluation settings. As shown in Fig. \ref{fig:coco_error}, under the strict evaluation metric IoU=0.75, overall AP (i.e., C75) of GCC-Net on the large and medium subset is 76.2\% and 77.2\%, respectively, outperforming the baseline method AutoAssign by a large margin (i.e., 4.2\% and 3.1\%, respectively), which demonstrates the effectiveness of our method in detecting more underwater objects. For the small subset, the $AP_{75}$ of our method is 46.8\%, while the AutoAssign is 45.6\%. This represents a notable improvement of 1.2\%, highlighting the capability of our method in effectively detecting small objects. And under the PASCAL IoU metric IoU=0.50, the $AP_{50}$ of our method on the small subset is 71.7\%, while the AutoAssign is 70.5\%. In the case of Loc (localization errors ignored), the AP of GCC-Net on the small subset is 72.3\%, exceeding the baseline method (i.e., 71.2\%) by 1.1\%, which demonstrates that the proposed method has a better regression ability than the baseline method. For the metric Oth, which removes all class confusion, we raise AP to 74.8\% on the small subset compared with 73.1\% of the baseline method AutoAssign. After all background and class confusion false positives are removed (metric BG), the proposed method achieves 99.3\% in AP performance on the small subset, surpassing the baseline method by 0.8\% (99.3\% vs 98.5\%) absolutely. The results indicate the strong detection capability of our method across various object scales, particularly for small objects. We believe the reason is the integration of the UIE model allows for clear visibility of objects in low-contrast underwater regions. 

\subsection{Qualitative Results}
Fig. \ref{fig:duo} and Fig. \ref{fig:others} show the qualitative results of sample images from the DUO, Brackish, TrashCan, and WPBB datasets. We can notice that the proposed method can deal with various challenges in UOD tasks, including different underwater environments, low-contrast and small objects, and densely-arranged objects. For example, as shown in Fig. \ref{fig:duo}, we present the ground truth, the enhanced images generated by water-MSR, and the detection results of GCC-Net on the DUO dataset.  As we can see, the enhanced images improve the visibility of small objects in low-contrast areas. The results clearly demonstrate the superior performance of our method in terms of dense small object localization and classification. We also present the qualitative results of GCC-Net on the Brackish, TrashCan, and WPBB datasets, shown in Fig. \ref{fig:others}. The promising results validate the versatility of our method across various water environments. However, there are still some failure cases, especially when the objects are lived heavily occluded. Future methods can focus on addressing these difficult cases.

\section{Conclusion}
In this paper, we propose an end-to-end gated cross-domain collaborative network (GCC-Net) to address the challenges of low-contrast and poor visibility in UOD tasks. Our method integrates the UIE and the UOD methods from a new perspective, mining and fusing the advantageous information between the two image domains. Our method is the first work to improve the performance of UOD from the perspective of cross-domain data interaction and fusion. The proposed GCC-Net comprises dedicated components, including an online UIE model ``water-MSR'', a cross-domain feature interaction module, and a gated feature fusion module. The proposed GCC-Net achieves state-of-the-art performance compared to recent GOD and UOD methods on the underwater datasets (DUO, Brackish, TrashCan, and WPBB datasets).  Furthermore, we hope this cross-domain collaborative paradigm will promote future work in underwater object detection and other multi-modal computer vision tasks.

\bibliographystyle{elsarticle-num} 
\bibliography{ref}

\begin{thebibliography}{10}
\expandafter\ifx\csname url\endcsname\relax
  \def\url#1{\texttt{#1}}\fi
\expandafter\ifx\csname urlprefix\endcsname\relax\def\urlprefix{URL }\fi
\expandafter\ifx\csname href\endcsname\relax
  \def\href#1#2{#2} \def\path#1{#1}\fi

\bibitem{li2021underwater}
C.~Li, S.~Anwar, et~al., Underwater image enhancement via medium
  transmission-guided multi-color space embedding, IEEE TIP 30 (2021)
  4985--5000.

\bibitem{chen2020reveal}
X.~Chen, Y.~Lu, et~al., Reveal of domain effect: How visual restoration
  contributes to object detection in aquatic scenes, arXiv:2003.01913 (2020).

\bibitem{dai2023edgeguided}
L.~Dai, H.~Liu, et~al., Edge-guided representation learning for underwater
  object detection (2023).
\newblock \href {http://arxiv.org/abs/2306.00440} {\path{arXiv:2306.00440}}.

\bibitem{islam2020fast}
M.~J. Islam, Y.~Xia, J.~Sattar, Fast underwater image enhancement for improved
  visual perception, IEEE RAL 5~(2) (2020) 3227--3234.

\bibitem{EfficientDets}
F.~Zocco, T.-C. Lin, et~al., Towards more efficient efficientdets and real-time
  marine debris detection, IEEE RAL 8~(4) (2023) 2134--2141.

\bibitem{liu2022twin}
R.~Liu, Z.~Jiang, et~al., Twin adversarial contrastive learning for underwater
  image enhancement and beyond, IEEE TIP 31 (2022) 4922--4936.

\bibitem{zhu2020autoassign}
B.~Zhu, J.~Wang, et~al., Autoassign: Differentiable label assignment for dense
  object detection, arXiv preprint arXiv:2007.03496 (2020).

\bibitem{qiao2021detectors}
S.~Qiao, L.-C. Chen, A.~Yuille, Detectors: Detecting objects with recursive
  feature pyramid and switchable atrous convolution, in: CVPR, 2021, pp.
  10213--10224.

\bibitem{dai2022ao2}
L.~Dai, H.~Liu, H.~Tang, Z.~Wu, P.~Song, Ao2-detr: Arbitrary-oriented object
  detection transformer, IEEE Transactions on Circuits and Systems for Video
  Technology (2022).

\bibitem{fu2022uncertainty}
Z.~Fu, W.~Wang, et~al., Uncertainty inspired underwater image enhancement, in:
  ECCV 2022, 2022, pp. 465--482.

\bibitem{dong2022underwater}
L.~Dong, W.~Zhang, W.~Xu, Underwater image enhancement via integrated rgb and
  lab color models, Signal Processing: Image Communication 104 (2022) 116684.

\bibitem{yeh2021lightweight}
C.-H. Yeh, C.-H. Lin, et~al., Lightweight deep neural network for joint
  learning of underwater object detection and color conversion, IEEE TNNLS
  33~(11) (2021) 6129--6143.

\bibitem{sun2022rethinking}
S.~Sun, W.~Ren, et~al., Rethinking image restoration for object detection 35
  (2022) 4461--4474.

\bibitem{land1964retinex}
E.~H. Land, The retinex, American Scientist 52~(2) (1964) 247--264.

\bibitem{tang2019efficient}
C.~Tang, U.~F. von Lukas, et~al., Efficient underwater image and video
  enhancement based on retinex, Signal, Image and Video Processing 13 (2019)
  1011--1018.

\bibitem{DUO}
C.~Liu, H.~Li, et~al., A dataset and benchmark of underwater object detection
  for robot picking, in: ICME Workshops, 2021, pp. 1--6.

\bibitem{brackish}
M.~Pedersen, J.~Bruslund~Haurum, R.~Gade, T.~B. Moeslund, Detection of marine
  animals in a new underwater dataset with varying visibility, in: CVPR
  Workshops, 2019, pp. 18--26.

\bibitem{hong2020trashcan}
J.~Hong, M.~Fulton, J.~Sattar, Trashcan: A semantically-segmented dataset
  towards visual detection of marine debris, arXiv preprint arXiv:2007.08097
  (2020).

\bibitem{ROIMIX}
W.~{Lin}, J.~{Zhong}, S.~{Liu}, T.~{Li}, G.~{Li}, Roimix: Proposal-fusion among
  multiple images for underwater object detection, in: ICASSP, 2020, pp.
  2588--2592.

\bibitem{huang2019faster}
H.~Huang, H.~Zhou, et~al., Faster r-cnn for marine organisms detection and
  recognition using data augmentation, Neurocomputing 337 (2019) 372--384.

\bibitem{liu2020towards}
H.~Liu, P.~Song, et~al., Towards domain generalization in underwater object
  detection, in: ICIP, 2020, pp. 1971--1975.

\bibitem{chen2023achieving}
Y.~Chen, P.~Song, et~al., Achieving domain generalization for underwater object
  detection by domain mixup and contrastive learning, Neurocomputing (2023).

\bibitem{UDD}
C.~Liu, Z.~Wang, et~al., A new dataset, poisson gan and aquanet for underwater
  object grabbing, IEEE TCSVT 32~(5) (2022) 2831--2844.

\bibitem{chen2022swipenet}
L.~Chen, F.~Zhou, S.~Wang, J.~Dong, N.~Li, H.~Ma, X.~Wang, H.~Zhou, Swipenet:
  Object detection in noisy underwater scenes, Pattern Recognition 132 (2022)
  108926.

\bibitem{song2023boosting}
P.~Song, P.~Li, L.~Dai, T.~Wang, Z.~Chen, Boosting r-cnn: Reweighting r-cnn
  samples by rpn’s error for underwater object detection, Neurocomputing 530
  (2023) 150--164.

\bibitem{tan2020efficientdet}
M.~Tan, R.~Pang, Q.~V. Le, Efficientdet: Scalable and efficient object
  detection, in: CVPR, 2020, pp. 10781--10790.

\bibitem{underwatermodel}
J.~Y. Chiang, Y.-C. Chen, Underwater image enhancement by wavelength
  compensation and dehazing, IEEE TIP 21~(4) (2012) 1756--1769.

\bibitem{watergan}
J.~Li, K.~A. Skinner, R.~M. Eustice, M.~Johnson-Roberson, Watergan:
  Unsupervised generative network to enable real-time color correction of
  monocular underwater images, IEEE RAL 3~(1) (2018) 387--394.

\bibitem{li2020underwater}
C.~Li, S.~Anwar, et~al., Underwater scene prior inspired deep underwater image
  and video enhancement, Pattern Recognition 98 (2020) 107038.

\bibitem{Sea-Thru}
D.~{Akkaynak}, T.~{Treibitz}, Sea-thru: A method for removing water from
  underwater images, in: CVPR, 2019, pp. 1682--1691.

\bibitem{USP}
C.~Li, S.~Anwar, F.~Porikli, Underwater scene prior inspired deep underwater
  image and video enhancement, Pattern Recognition 98 (2020) 107038.

\bibitem{liu2021swin}
Z.~Liu, Y.~Lin, Y.~Cao, H.~Hu, Y.~Wei, Z.~Zhang, S.~Lin, B.~Guo, Swin
  transformer: Hierarchical vision transformer using shifted windows, in: ICCV,
  2021, pp. 10012--10022.

\bibitem{jobson1997multiscale}
D.~J. Jobson, Z.-u. Rahman, G.~A. Woodell, A multiscale retinex for bridging
  the gap between color images and the human observation of scenes, IEEE TIP
  6~(7) (1997) 965--976.

\bibitem{FasterRCNN}
S.~{Ren}, K.~{He}, et~al., Faster r-cnn: Towards real-time object detection
  with region proposal networks, IEEE TPAMI 39~(6) (2017) 1137--1149.

\bibitem{deformable}
X.~Zhu, W.~Su, et~al., Deformable detr: Deformable transformers for end-to-end
  object detection, in: ICLR, 2020.

\bibitem{mmdetection}
K.~Chen, J.~Wang, et~al., {MMDetection}: Open mmlab detection toolbox and
  benchmark, arXiv preprint arXiv:1906.07155 (2019).

\bibitem{loshchilov2017decoupled}
I.~Loshchilov, F.~Hutter, Decoupled weight decay regularization, arXiv preprint
  arXiv:1711.05101 (2017).

\bibitem{CascadeRCNN}
Z.~{Cai}, N.~{Vasconcelos}, Cascade r-cnn: Delving into high quality object
  detection, in: CVPR, 2018, pp. 6154--6162.

\bibitem{sabl}
J.~Wang, W.~Zhang, et~al., Side-aware boundary localization for more precise
  object detection, in: ECCV, 2020, pp. 403--419.

\bibitem{gfl}
X.~Li, W.~Wang, et~al., Generalized focal loss: Learning qualified and
  distributed bounding boxes for dense object detection 33 (2020) 21002--21012.

\bibitem{wang2022yolov7}
C.-Y. Wang, A.~Bochkovskiy, et~al., Yolov7: Trainable bag-of-freebies sets new
  state-of-the-art for real-time object detectors, arXiv preprint
  arXiv:2207.02696 (2022).

\bibitem{liang2022excavating}
X.~Liang, P.~Song, Excavating roi attention for underwater object detection,
  in: ICIP, IEEE, 2022, pp. 2651--2655.

\bibitem{zhang2022underwater}
W.~Zhang, P.~Zhuang, et~al., Underwater image enhancement via minimal color
  loss and locally adaptive contrast enhancement, IEEE TIP 31 (2022)
  3997--4010.

\bibitem{selvaraju2017grad}
R.~R. Selvaraju, M.~Cogswell, et~al., Grad-cam: Visual explanations from deep
  networks via gradient-based localization, in: ICCV, 2017, pp. 618--626.

\bibitem{lin2014microsoft}
T.-Y. Lin, M.~Maire, Microsoft coco: Common objects in context, in: ECCV, 2014,
  pp. 740--755.

\end{thebibliography}

\end{document}